\documentclass[10pt,twocolumn,letterpaper]{article}

\usepackage[pagenumbers]{wacv} 

\usepackage{graphicx}
\usepackage{amsmath}
\usepackage{amssymb}
\usepackage{booktabs}

\usepackage{times}
\usepackage{epsfig}
\usepackage{graphicx}
\usepackage{amsmath}
\usepackage{amssymb}
\usepackage{booktabs}
\usepackage{standalone}
\usepackage{siunitx}
\usepackage[most]{tcolorbox}
\usepackage{multirow}
\usepackage{array}
\usepackage{xinttools}

\usepackage[pagebackref,breaklinks,colorlinks]{hyperref}

\newlength{\imageheight}

\usepackage[capitalize]{cleveref}
\crefname{section}{Sec.}{Secs.}
\Crefname{section}{Section}{Sections}
\Crefname{table}{Table}{Tables}
\crefname{table}{Tab.}{Tabs.}

\def\wacvPaperID{1949} 
\def\confName{WACV}
\def\confYear{2024}

\newcommand{\cbar}[1]{            
            \begin{tabular}{lr}
            \multicolumn{2}{c}{\reflectbox{\includegraphics[width=#1]{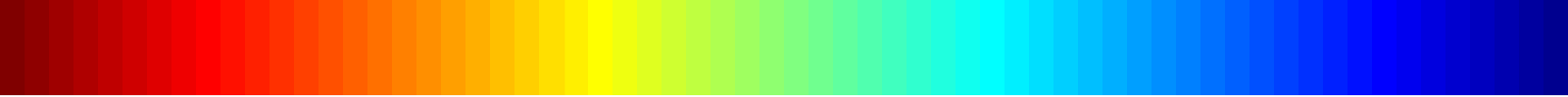}}} \\[2pt]
            $0^{\circ}$ & $60^{\circ}$
            \end{tabular}}

\newcommand{\note}[1] {\hspace{0pt}} 

\begin{document}

\title{FOUND: \underline{F}oot \underline{O}ptimization with \underline{U}ncertain \underline{N}ormals for Surface \underline{D}eformation Using Synthetic Data}

\author{
  \mbox{Oliver Boyne \qquad Gwangbin Bae \qquad James Charles \qquad Roberto Cipolla} \\
  Department of Engineering, University of Cambridge, U.K. \\
  \tt{\{ob312, gb585, jjc75, rc10001\}@cam.ac.uk} \\
}

\maketitle
\vspace{-1pt}
\begin{abstract}
  \vspace{-2pt}
  Surface reconstruction from multi-view images is a challenging task, with solutions often requiring a large number of sampled images with high overlap. We seek to develop a method for few-view reconstruction, for the case of the human foot. To solve this task, we must extract rich geometric cues from RGB images, before carefully fusing them into a final 3D object. Our FOUND approach tackles this, with 4 main contributions: (i) \textit{SynFoot}, a synthetic dataset of 50,000 photorealistic foot images, paired with ground truth surface normals and keypoints; (ii) an uncertainty-aware surface normal predictor trained on our synthetic dataset; (iii) an optimization scheme for fitting a generative foot model to a series of images; and (iv) a benchmark dataset of calibrated images and high resolution ground truth geometry. We show that our normal predictor outperforms all off-the-shelf equivalents significantly on real images, and our optimization scheme outperforms state-of-the-art photogrammetry pipelines, especially for a few-view setting. We release our synthetic dataset and baseline 3D scans to the research community.
\end{abstract}

\section{Introduction}

\begin{figure}
  \centering
  \includegraphics[width=\columnwidth]{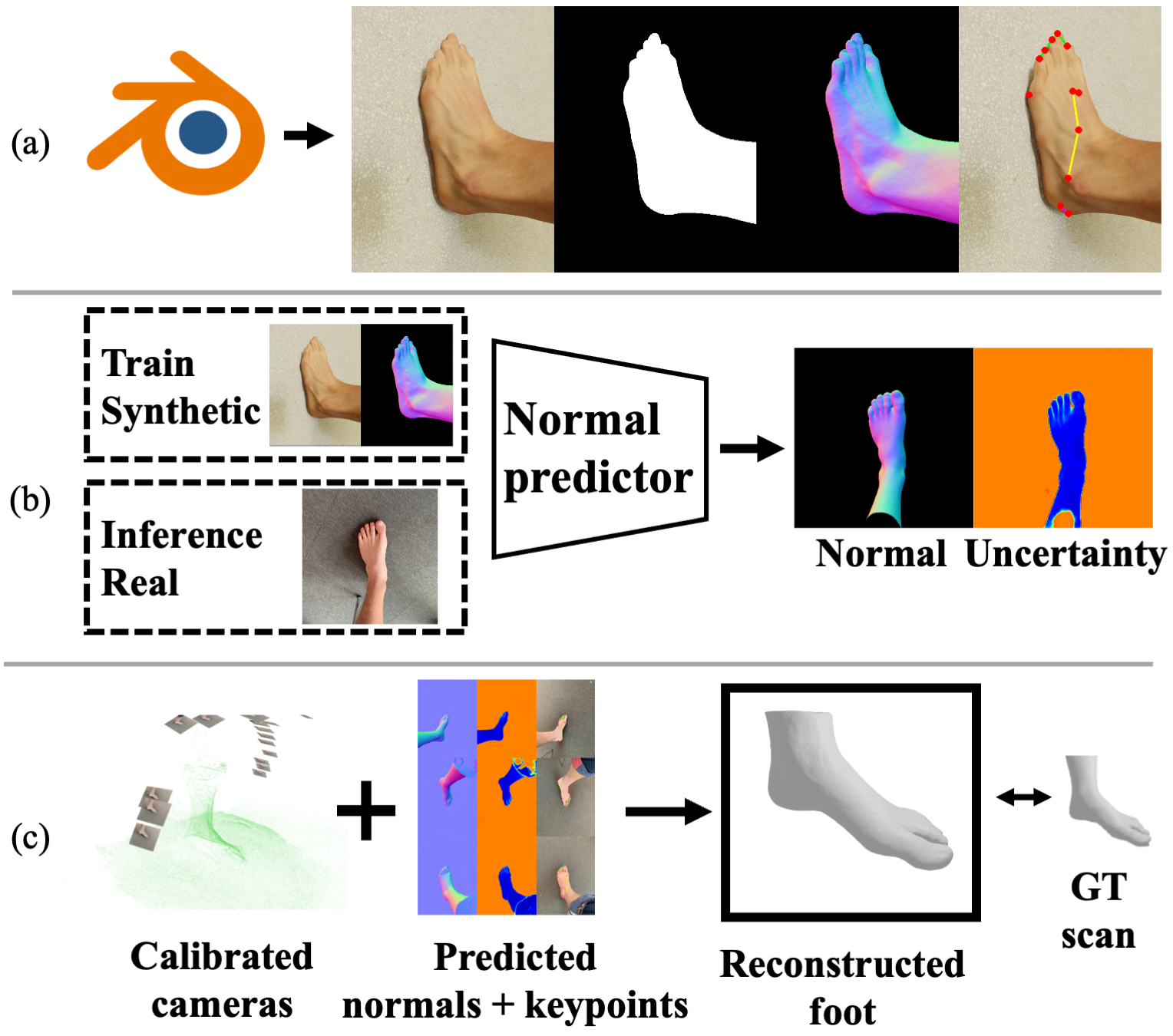}
  \caption{Method overview: (a) we use Blender \cite{blender} to synthetically render foot images, masks, surface normals, and keypoints; (b) we train a normal predictor on this data; (c) we predict normals on real images and optimize in a multi-view, calibrated setup to reconstruct the foot, evaluating on a ground truth scan.}
  \label{fig:splash}
\end{figure}

3D reconstruction of human body parts from images is a challenging computer vision task, of significant interest to the health, fashion and fitness industry. In this paper, we address the problem of human foot reconstruction. Shoe retail, orthotics, and personal health monitoring all benefit from accurate models of the foot, and the growth of the digital market for these industries has made the prospect of recovering a 3D foot model from images very appealing.

Existing solutions for foot reconstruction fit into one of four categories:
(i) expensive scanning equipment \cite{Volumental, yeti3d, ArtecLeo}; 
(ii) reconstruction of noisy point clouds, \eg from depth maps \cite{lunscher2017point} or phone-based sensors such as a \textit{TrueDepth} camera \cite{Xesto}; 
(iii) Structure from Motion (SfM) followed by Multi-View Stereo (MVS) \cite{SFM_2016_COLMAP, MVS_2016_COLMAP};
and (iv) fitting generative foot models to image silhouettes \cite{kok2020footnet}.

We find none of these solutions are satisfactory for accurate scanning in a home setting: expensive scanning equipment is not accessible to most consumers; phone-based sensors are limited in availability and ease of use, and noisy point clouds are difficult to use for downstream tasks such as rendering and taking measurements; SfM is dependent on a large number of input views to match dense features between images, and MVS can also produce noisy point clouds; and (until recently) foot generative models have been low quality and restrictive, and using only silhouettes from images limits the amount of geometrical information that can be obtained from the images, especially problematic in a few-view setting. The performance of these methods is also limited by the lack of paired images and 3D ground truth data for feet available for training.\\

To this end, we introduce FOUND, \textbf{F}oot \textbf{O}ptimisation using \textbf{U}ncertain \textbf{N}ormals for surface \textbf{D}eformation - which seeks to improve on typical multi-view reconstruction optimization schemes by leveraging per-pixel surface normals along with uncertainties. Similar to \cite{kok2020footnet}, our method only requires a small number of calibrated RGB images as input. While \cite{kok2020footnet} relied only on silhouettes (which lack geometric information), we exploit additional cues - surface normals and keypoints. To address the lack of data, we also release a large-scale dataset of photorealistic synthetic images paired with ground truth labels for such cues. We outline our key contributions as follows: 

\begin{itemize}
\itemsep0em 
\item To facilitate research on 3D foot reconstruction, we release \textit{SynFoot}, a \textbf{large-scale synthetic dataset} of 50,000 photorealistic foot images, coupled with accurate silhouette, surface normal, and keypoint labels. While collecting such data for real images requires expensive scanning equipment, our dataset is highly scalable. Despite only containing 8 real world foot scans, we show that our synthetic dataset sufficiently captures variation within foot images enough for downstream tasks to generalize to real images. We also release an \textbf{evaluative dataset} of 474 images of 14 real feet, each paired with ground truth per-pixel surface normals and a high-resolution 3D scan. Finally, we release our custom Python library for generating large scale synthetic datasets efficiently using Blender \cite{blender}.

\item We demonstrate that an \textbf{uncertainty-aware surface normal estimation network}, trained purely on our synthetic data from 8 foot scans, generalizes to real in-the-wild foot images. We use aggressive appearance and perspective augmentation to close the domain gap between synthetic and real foot images. The network estimates per-pixel surface normals and corresponding uncertainty. The uncertainty is useful in two regards: we can get accurate silhouettes for free (\ie without training a separate network) by thresholding the uncertainty; and we can use the estimated uncertainty to weight the surface normal loss in our optimization scheme, increasing robustness against potential inaccuracy of the predictions made in certain views.

\item We propose an \textbf{optimization scheme} capable of fitting a generative foot model \cite{boyne2022find} to a set of calibrated images with predicted surface normals and keypoints via differentiable rendering. Our pipeline is uncertainty-aware, capable of reconstructing a watertight mesh from very few views, can be applied to data collected from a consumer mobile phone, and beats state-of-the-art photogrammetry for surface reconstruction.

\end{itemize}

\section{Related work}

\paragraph{Synthetic dataset generation.} In recent years, the capabilities for rendering photorealistic images in large quantities has improved significantly. This has resulted in a growing interest in generating high quality synthetic datasets for computer vision tasks.
Synthetic data has the benefits of being much more scalable, cheap and accurate than typical data collection pipelines - especially for tasks difficult or impossible for human labellers. Many works have shown that synthetic data can be used as the primary source of training data for various deep learning problems. For human body reconstruction tasks, synthetic data has been used for bodies \cite{Varol_2017_CVPR}, faces \cite{bae2023digiface, wood2021fake}, and eyes \cite{wood2015rendering}.

\paragraph{Single-image surface normal estimation.} The goal of this problem is to estimate the pixel-wise surface normal, defined in the camera reference frame. While classical approaches \cite{SNfromRGB_2007_Hoiem, SNfromRGB_2014_Ladicky} relied on handcrafted image features and simplified the problem by discretizing the output and solving a classification problem, recent methods \cite{SNfromRGB_2019_SR, SNfromRGB_2020_TiltedSN, SNfromRGB_2021_EESNU} use deep neural networks to directly regress the per-pixel normal. Unlike depth, surface normals are not affected by scale ambiguity and can be estimated from low-level cues like edges and shading \cite{SNfromRGB_2021_EESNU}, improving generalization. We follow the approach by Bae \etal \cite{SNfromRGB_2021_EESNU} to quantify the uncertainty associated with the normal predictions and use it to improve the robustness of our foot optimization.

\paragraph{Normal integration.} A surface normal map can be integrated to recover 3D shape up to a scale ambiguity. Methods like \cite{cao2021normal, xu2022bilateral} optimize per-pixel depth such that the normal map computed from the depth map is consistent with the input normal map. However, such methods (1) cannot recover the scale, (2) are not applicable for multi-view images, (3) are unable to reconstruct unseen parts of the object and (4) are not uncertainty-aware (i.e. sensitive to inaccurate normal predictions). Conversely, we optimize the foot shape based on multi-view uncertainty-aware normal and keypoint constraints, so that the contribution of high-uncertainty predictions can be down-weighted. The optimized foot shape is both water-tight and accurately scaled, useful for commercial applications like virtual shoe try-on.

\paragraph{Human foot reconstruction. } Human foot reconstruction is a task of interest to the footwear and orthotics industries.

Some foot reconstruction methods seek to reconstruct point clouds of feet \cite{lunscher2017point}, often leveraging LiDAR or structured light sensors available in some modern mobile phones. These sensors are not ubiquitous however, and these often noisy point cloud reconstructions are limited in their use for downstream tasks, often useful only to take a small number of measurements of the foot.

Other methods fit a generative model to predicted image silhouettes \cite{kok2020footnet}. Building statistically-derived generative models for feet is difficult, in part because there are no large scale datasets available to the research community, outside of population measurement statistics \cite{jurca2011dorothy, jurca2019analysis}. While generative models of the foot that use Principal Component Analysis (PCA) are not new \cite{amstutz2008pca}, these have been very limited in their resolution and capabilities until recently.
 Recent work by Osman \etal{} \cite{SUPR:2022} introduced SUPR, a PCA model of the human foot to be combined with the SMPL \cite{SMPL:2015} full body model for the task of expressive, full body reconstruction.
Boyne \etal \cite{boyne2022find} also recently released a generative model of the human foot, FIND. Instead of PCA, this model uses an implicit network to define per-vertex deformations to deform a template mesh into a target pose and shape. A dataset of high resolution foot scans was released with FIND.

\paragraph{Multi-view reconstruction. } Reconstruction of depth and geometry from images requires known camera positions and internal parameters. These can be obtained directly from a capturing device using Augmented Reality (AR) technologies and Inertial Measurement Units (IMUs), or via a sparse 3D reconstruction from Structure From Motion (SfM) \cite{SFM_2016_COLMAP}.

From this, a common method for reconstructing surface geometry is Multi-View Stereo (MVS) \cite{MVS_2016_COLMAP, campbell2008using}, which involves the fusion of features matched across views. To produce a mesh as output, a surface reconstruction algorithm is required \cite{kazhdan2006poisson, bernardini1999ball}.

Recent work has ventured into using neural rendering for surface reconstructions \cite{mildenhall2021nerf, yariv2020multiview}, in which some neural representation of a 3D scene is learned which matches the reference views when rendered. Such methods are often dependent on a high number of input views and require large amounts of training time to produce a reconstruction.

\section{Method}
\label{sec:method}

\subsection{SynFoot - Synthetic dataset}

\begin{figure}
   \centering
   \includegraphics[width=\columnwidth]{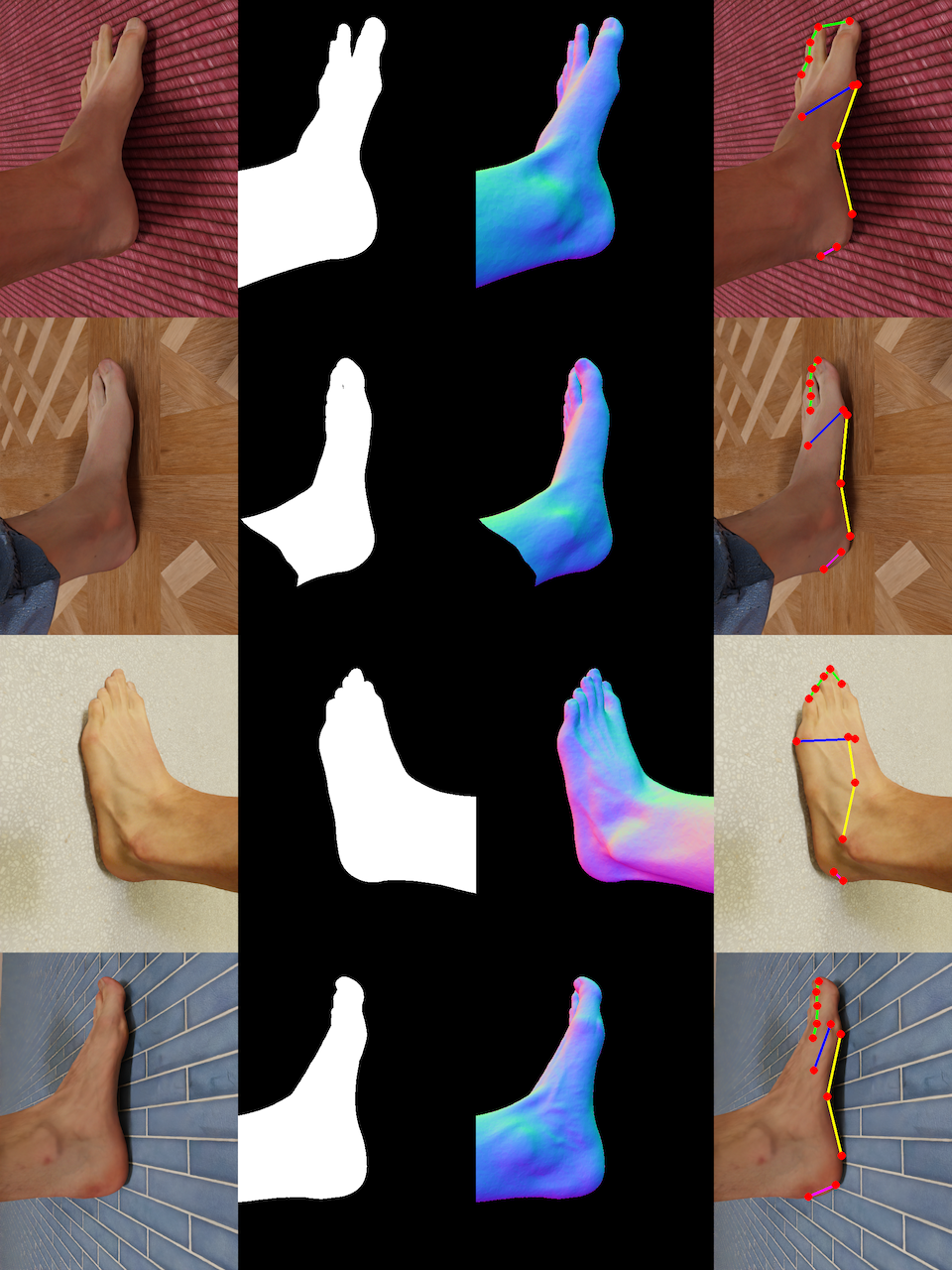}
   \caption{Samples from \textit{SynFoot}, our synthetic dataset. We show (a) RGB, (b) silhouettes, (c) surface normals, and (d) keypoints. Further examples included in the supplementary.}
   \label{fig:synth_dset}
\end{figure}

\vspace{-3pt}
We generate 50,000 images at 480 $\times$ 640 resolution using Blender's \cite{blender} Cycles engine. We replicate the behavior of images typically used for foot reconstruction commercial applications: captured from a handheld device, in a vertical arc around the foot, with the foot placed on the floor. We generate this dataset using a custom Python library built for large scale synthetic dataset rendering, which we release to the research community.

\vspace{-6pt}

\paragraph{Camera.} We sample camera positions in a vertical arc above the foot, uniformly sampling a radius from \SI{30}{} to \SI{40}{\cm}, angle to the vertical from - 0.4$\pi$ to 0.4$\pi$\,$\textrm{rad}$, and lateral displacement from \SI{0}{} to \SI{10}{\cm}. We use a camera focal length of \SI{30}{\mm}, and point the camera at the world origin.

\paragraph{Leg models.} We find current generative models lack necessary photorealism, and do not capture enough geometry up the leg, for use in our synthetic dataset. As a result, we use 8 foot and leg scans adapted from the Foot3D dataset \cite{boyne2022find} - we find that a small number of leg scans, coupled with a large variety in environment and viewpoint, provide sufficient training data to train networks on downstream tasks. We also align one of two trouser models (with a randomly selected texture) to the model.

As the scan data does not capture the entire leg, there exists a horizontal cutoff plane on each mesh. From most views this is not visible, but certain viewpoints would suffer reduced realism as a result. We avoid this by (a) covering the plane with trousers, and (b) avoiding viewpoints where we detect that the cutoff plane would occupy more than 20\% of the rendered image.

\paragraph{Lighting.} We sample one of 14 HDRI environments \cite{polyhaven}, with a random intensity value to simulate variable lighting conditions. We add a point light at a random horizontal position, to add further realistic lighting variation.
We also add a plane behind the camera, with dimensions of a typical mobile phone, to replicate the shadow typically seen in mobile phone captured images.

\paragraph{Floor surface.} We sample one of 34 textures collected from a number of license-free asset stores \cite{poliigon, polyhaven, ambientcg, cgbookcase} to model variety in flooring. We use diffuse, normal, reflectance, and roughness maps where available to provide realism to the surface. To replicate specular reflections seen on real surfaces, we apply a uniformly sampled [0, 0.3] specularity value to 20\% of the dataset samples.

\paragraph{Compositing. } As shown in Figure \ref{fig:synth_dset}, for each sample we render: \textbf{RGB} images; \textbf{silhouettes} of the visible leg scan; \textbf{surface normals} relative to the camera, normalized from [-1, 1] to [0, 255], with XYZ corresponding to RGB; and \textbf{keypoints} based on 12 handlabelled keypoints on each scan (detailed in Figure \ref{fig:kp_labelling}).

\subsection{Label prediction}

So that we can optimize a generative foot model to real images, we seek to use our synthetic dataset to learn to predict useful representations of real data for subsequent fitting: per-pixel surface normals, silhouettes, and keypoints.

\vspace{-2pt}

\paragraph{Surface normal prediction.} Following \cite{SNfromRGB_2021_EESNU}, we predict the surface normal \textit{probability distribution} and train the network by minimizing the negative log-likelihood (NLL) of the ground truth. Given $N$ pixels (indexed with $i$) with ground truth surface normal $\mathbf{n}_i^\text{gt}$, the loss can be written as

\begin{equation}   
\begin{aligned}
\label{eqn:normal-nll}
\mathcal{L}_{\text{norm}}^\textrm{train} 
&= - \frac{1}{N} \sum^N_i 
\log p_{\text{AngMF},i}(\mathbf{n}^\text{gt}_i|\boldsymbol{\mu}_i,\kappa_i) \\
&= \frac{1}{N} \sum^N_i 
\kappa_i \cos^{-1} \boldsymbol{\mu}_i^T \mathbf{n}_i^\text{gt}
+ \log \frac{(1 + \exp(-\kappa_i \pi))}{(\kappa^2_{i} +1)},
\end{aligned}
\end{equation}

\noindent
where $\boldsymbol{\mu}_i$ and $\kappa_i$ are the predicted mean direction and the concentration parameter. The objective of the training is to minimize the angle between $\boldsymbol{\mu}_i$ and $\mathbf{n}_i^\text{gt}$ while reducing $\kappa_i$ for the pixels with high error. The expected angular error can then be used to quantify the uncertainty \cite{SNfromRGB_2021_EESNU}.

For our synthetic dataset, the ground truth is provided only for the foot. To encourage the network to estimate high uncertainty for the background, we generate noisy labels for the background pixels by sampling them uniformly on a unit hemisphere facing the camera. For the background pixels, the same loss (Equation \ref{eqn:normal-nll}) is computed and is down-weighted by a factor $0.1$. When back-propagating the loss for the background pixels, the gradient is detached for $\boldsymbol{\mu}$ to only influence the uncertainty.

\paragraph{Data augmentation.} While our synthetic images look realistic, there still remains substantial domain gap with respect to the in-the-wild real images of human foot. Inspired by \cite{bae2023digiface}, we add aggressive data augmentation to close this domain gap. Specifically, we add random horizontal flipping, JPEG compression and Gaussian blur/noise, together with randomization of brightness, contrast, saturation and hue. We also add perspective augmentation by rotating the camera around the yaw, pitch and roll axes. More detail is provided in the supplementary material.

\vspace{-2pt}

\paragraph{Silhouette prediction.} From the surface normal prediction, we can get the expected value of the angular error. We conveniently obtain silhouettes by masking out the pixels with uncertainty higher than $30^{\circ}$.

\vspace{-2pt}

\begin{figure}
    \centering
    \includegraphics[width=\columnwidth]{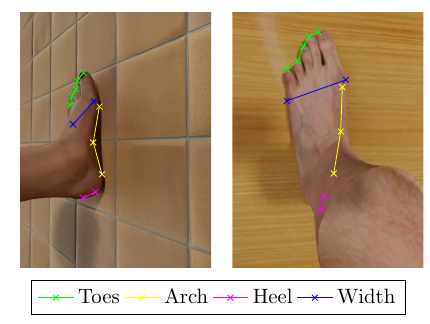}
    \caption{12 keypoints defined for our dataset on two synthetic images - detailed keypoint definitions included in the supplementary.}
    \label{fig:kp_labelling}
\end{figure}

\newcommand{\loss}[1]{\mathcal{L}_{\textrm{#1}}}
\newcommand{\gt}[1]{#1^{\textrm{gt}}}
\newcommand{\norm}[1]{\left\lVert#1\right\rVert}
\def\l2{\ell_2}

\paragraph{Keypoint prediction.} Similar to surface normal prediction, we estimate the probability distribution for the keypoint locations, parameterized with a 2D Gaussian with a diagonal covariance matrix. We train a ResNet-50 \cite{he2016deep} model to  predict for an image, for each of the 12 keypoints visualized in Figure \ref{fig:kp_labelling}: its 2D image position $\mathbf{k}$ normalized to unit image coordinates $\bar{\mathbf{k}}$; a visibility flag $v$ (whether within the image bounds); and predicted uncertainty $\boldsymbol{\sigma}=(\sigma_x, \sigma_y)$. We learn to predict keypoints and uncertainty by comparing against ground truth keypoints $\mathbf{\gt{k}}$,
\vspace{-10pt}

\begin{equation}
\begin{split}
 \loss{kp}^\textrm{train} = \frac{1}{N} \sum_i^N \hat{v}_{i} \bigg( & \norm{\left( 
\begin{matrix}
(\mathbf{\bar k}_{i,j,x} -  {\mathbf{\gt{\bar{k}}}}_{i,j,x} )/\bar \sigma_{x,i}
\\ 
(\mathbf{\bar k}_{i,j,y} -  {\mathbf{\gt{\bar{k}}}}_{i,j,y} )/\bar \sigma_{y,i}
\end{matrix}
\right)}_2^2\\ 
 & + \log{\bar{\sigma}_{x,i}^2 \bar{\sigma}_{y,i}^2} \bigg)
\end{split}
   \end{equation}

We learn visibilities with an $\ell_2$ loss comparing to ground truth visibility.
Similarly to our normal network, we train on our 50,000 synthetic images, adding affine and pixel color augmentations to the input data to improve the quality of training. We include examples showing the performance of the keypoint predictor on real images in the supplementary.

\subsection{Multi-view fusion}

\begin{figure*}
    \centering
    \centering

    \begingroup

    \setlength{\tabcolsep}{2pt} 
    \newcolumntype{C}[1]{>{\centering\let\newline\\\arraybackslash\hspace{0pt}}m{#1}}

    \def\imH{0.3}

\settoheight{\imageheight}{%
  \includegraphics[width=\imH \textwidth, keepaspectratio]{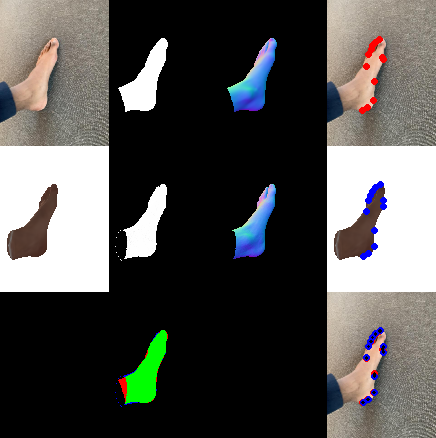}%
}

    \newcommand{\recheaders}{\begin{tabular}{*4{C{0.07\textwidth}}}
    (a) & (b) & (c) & (d)
    \end{tabular} }

    \newcommand{\recrows}{ \begin{tabular}{c}
    (i) \\[0.3\imageheight] (ii) \\[0.3\imageheight] (iii) 
    \end{tabular} }

    \begin{tabular}{p{0.4cm}ccc}
    & \recheaders & \recheaders  & \recheaders\\
    \recrows & \raisebox{-.5\height}{\includegraphics[width=\imH \textwidth]{images/reconstr_vis/0038_exp_07_03_23/views/stage_01/view_07_crop.png}} &
    \raisebox{-.5\height}{\includegraphics[width=\imH \textwidth]{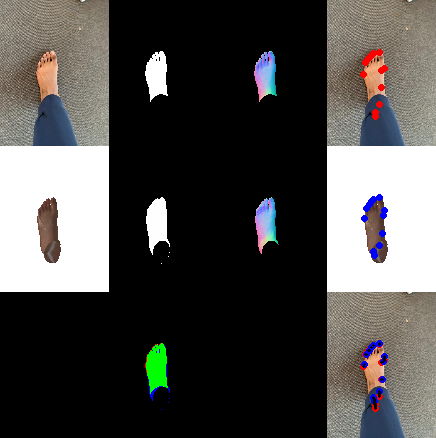}}&
    \raisebox{-.5\height}{\includegraphics[width=\imH \textwidth]{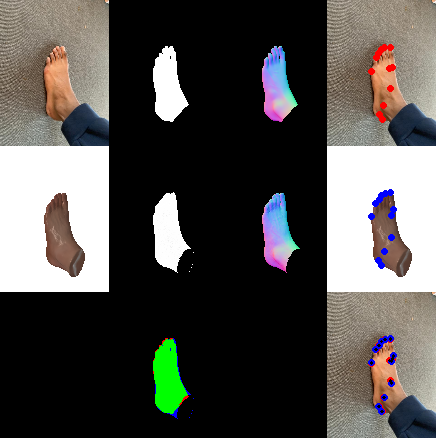}}
        
    \end{tabular}
    \endgroup
    \vspace{1.5mm}
    \caption{Here, we show our method optimizing to input images of a real foot, visualizing for three views: (a) RGB, (b) silhouette, (c) normals, and (d) keypoints, for (i) real input image, (ii) fitted FIND model, and (iii) the error. Note that the RGB for FIND uses FIND's default texture. To view the reconstruction quality compared to COLMAP and the GT scan, see Figure \ref{fig:reconstr}.}
    \label{fig:vis_reconstr}
\end{figure*}

Once we can predict surface normals, silhouettes and keypoints for real input images, we seek to fit a generative foot model to the images.

\paragraph{Generative model.} For few-view fitting, a generative model is critical to enforce a strong prior over the foot space, and reduce the dimensionality of the optimization task. We use FIND \cite{boyne2022find}, an implicit-explicit model which captures shape, pose and texture. 
The model is composed of a template mesh and an MLP $F$. The MLP predicts, for given shape and pose encodings $z_s$ and $z_p$, the deformation for a template vertex at position $x$ to deform it to the target shape,

\begin{equation}
    \Delta x = F(x, z_s, z_p)
\end{equation}

The application of $F$ to every point $x$ on the template results in a deformed, explicit mesh. This is then  transformed by registration parameters $r$, $t$, and $s$ - Euler rotation, translation, and XYZ scaling respectively. The FIND model is also capable of producing per-vertex colours parameterized by a texture encoding - in this work, we fix this encoding, as our focus is on accurate geometry reconstruction.

\paragraph{Differentiable rendering.}
Using PyTorch3D \cite{ravi2020pytorch3d}, we differentiably render surface normals and silhouettes in the camera reference frame, and project keypoints onto the image, for optimization.

\subsection{Optimization}

We now have the tools to optimize the FIND model to the predicted labels. We optimize the registration (alignment) parameters $\{r, t, s\}$ and shape and pose parameters $\{z_p, z_s\}$. We optimize over $N$ views, each of size ($H$, $W$).

\paragraph{Keypoint loss. } For $K$ keypoints, we compute the $\l2$ distance between the $i$th keypoint projected in the $j$th view and normalized to unit coordinates $\mathbf{\bar k}_{i,j}$ and that predicted by our keypoint predictor $\mathbf{\hat{\bar k}}_{i,j}$. We ignore `invisible' keypoints using the visibility flag $\hat{v}$, and use our predicted normalized uncertainty $(\bar\sigma_{x,i}, \bar\sigma_{y,i})$,

      \begin{equation}
      \small
      \loss{kp}^\textrm{opt} = \frac{1}{N K} \sum_i^N \sum_j^K \hat{v}_{i,j} \norm{ \left( \begin{matrix}
(\mathbf{\bar k}_{i,j,x} -  {\mathbf{\hat{\bar{k}}}}_{i,j,x} )/\bar \sigma_{x,i}
\\ 
(\mathbf{\bar k}_{i,j,y} -  {\mathbf{\hat{\bar{k}}}}_{i,j,y} )/\bar\sigma_{y,i}
\end{matrix}
\right)}_2^2
   \end{equation}

We use $\sigma$ to down-weight the error for keypoints with high uncertainty - this typically corresponds to keypoints occluded in the image.

\paragraph{Normal loss. } Given the estimated per-pixel surface normal probability distribution, parameterized by $\boldsymbol{\mu}$ and $\kappa$, we minimize the negative log-likelihood of the surface normal of the rendered foot model. The loss can be written as,

\begin{equation}
\label{eqn:normal-fitting}
\loss{norm}^\textrm{opt} = 
\frac{1}{N} \sum_i^N 
\kappa_i \cos^{-1} \boldsymbol{\mu}_i^T \mathbf{n}_i^\text{render}.
\end{equation}

Equation \ref{eqn:normal-fitting} differs from \ref{eqn:normal-nll} in that the term only dependent on $\kappa$ is discarded as it is not being updated. Equation \ref{eqn:normal-fitting} reduces the angular error between the predicted normal and the current estimate of the foot shape, while down-weighting the error for predictions with high uncertainty (i.e. low $\kappa$).

\paragraph{Silhouette loss.} An $\ell_2$ loss between the rendered FIND model silhouette, and the pseudo-GT silhouette (found by thresholding the predicted normal uncertainty at $30^{\circ}$).

\subsection{Implementation details}

\paragraph{Label prediction.} We train the keypoint and surface normal predictors on our synthetic dataset. We use the Adam optimizer \cite{kingma2014adam} for both, at a learning rate of 0.0001. We train the normal predictor for 20 epochs, and the keypoint predictor for 70.

\paragraph{Optimization scheme.} We first downsample our images to increase speed and reduce memory usage.
We optimize in two stages: (i) \textit{registration}, with only FIND's $\{r, s, t\}$ as free parameters, and only a keypoint loss; and (ii) \textit{deformation}, with free parameters $\{r, s, t, z_s, z_p\}$, and keypoint, silhouette and normal losses present.

We use the Adam optimizer, with a learning rate of 0.001, for 250 epochs and 1000 epochs respectively for the two stages. On a Titan X GPU, this takes approximately 2 minutes for 3 input views. We show in Figure \ref{fig:vis_reconstr} a visualization of the result of a reconstruction.

Where we restrict views for experimentation, we choose the restricted views by selecting views distributed evenly in the $y$ direction (left-to-right as the images were taken).

\paragraph{Evaluative dataset.} As described in Section \ref{sec:exp}, we use camera extrinsic and intrinsic parameters obtained using Structure from Motion in COLMAP \cite{SFM_2016_COLMAP}. This method requires a certain amount of overlap between views. We use these cameras to provide a fair comparison between our method and COLMAP, only measuring the reconstruction performance and not the quality of the camera calibration, and so that we can evaluate on our ground truth dataset.

In practice, these parameters could be obtained directly by using tracking systems on mobile devices (\eg ARKit on iOS). We show in Section \ref{sec:future_work} that our method can reconstruct successfully with cameras obtained via such approaches.

\subsection{Experiments}

\begin{figure}
    \centering
    \includegraphics[width=\columnwidth]{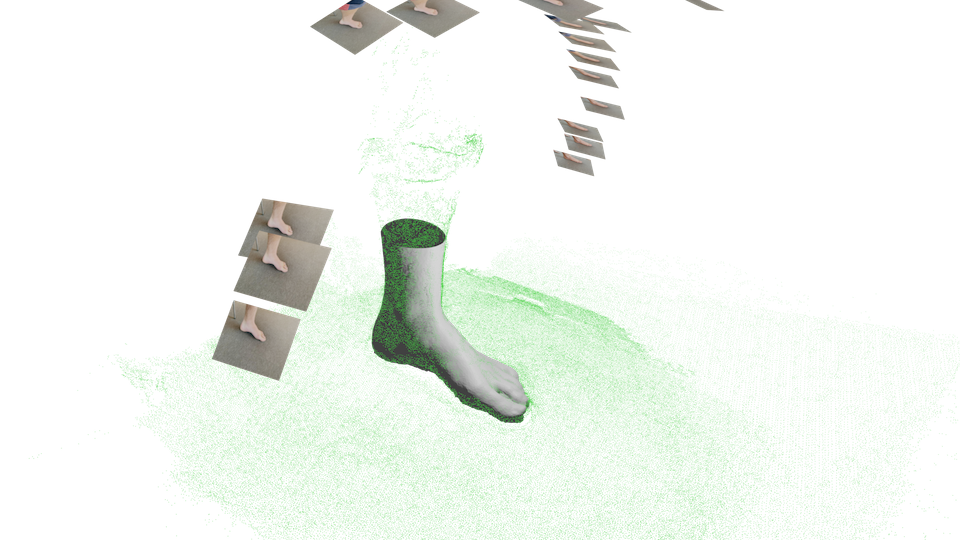}
    \caption{An example of a scan from our dataset. We show a sample of the calibrated images; the ground truth scan in grey; and the COLMAP reconstructed dense point cloud in green.}
    \label{fig:scan}
\end{figure}

\label{sec:exp}

\paragraph{Dataset.} To evaluate multi-view reconstruction performance, we collect a dataset of calibrated image captures around static feet, alongsde 3D ground truth scans from an Artec Leo \cite{ArtecLeo} scanner.

We perform sparse reconstruction via COLMAP \cite{SFM_2016_COLMAP}, sufficient for relative camera alignment, with a scale ambiguity. 
While this would be sufficient for shape fitting (albeit with a scale ambiguity), to be able to evaluate our normal predictor and our fitting process, it is necessary to align the COLMAP result to the ground truth scan.

To do this, we perform dense reconstruction in COLMAP \cite{MVS_2016_COLMAP}. We run outlier detection on this reconstructed dense point cloud, and floor detection and removal \cite{Zhou2018} on both the Leo scan and the COLMAP point cloud. We rotate and translate both so that the (now removed) floors are horizontal and at a height of zero. To align the two precisely, we run an optimization scheme to minimize the chamfer loss between the dense point cloud and the 3D scan, optimizing over four parameters: in-plane rotation, in-plane translation, and isotropic scale. We also use some outlier rejection stages to improve the accuracy of this process. An example output of this process can be seen in Figure \ref{fig:scan}.

Once this process is complete, we also conveniently can render the ground truth meshes onto each captured image to get ground truth surface normals, examples of which are in  Figure \ref{fig:normal_qual}. Note that this process was incredibly labour intensive, and produces a far smaller number of paired RGB and surface normal images than in our synthetic dataset.

The completed dataset contains 14 scenes of scanned feet, with a total of 474 calibrated images.

\vspace{-3pt}

\paragraph{Normal predictor evaluation protocol.} The accuracy of our surface normal estimation network is compared against COLMAP \cite{MVS_2016_COLMAP} and Bae \etal \cite{SNfromRGB_2021_EESNU}. 20 images are sampled from each scan to form a test set of 280 images. Pixel-wise angular error is computed and the mean, median and root-mean-squared errors are reported. We also report the percentage of pixels with error less than $[11.25^{\circ}, 22.5^{\circ}, 30^{\circ}]$. As we focus on the surface reconstruction of the foot, we discard the prediction for legs and trousers by slicing the ground truth mesh at 20 cm above the floor, and evaluating only on the pixels below the cutoff.

\vspace{-8pt}

\paragraph{Optimization evaluation protocol.} We evaluate the quality of our 3D optimization method by comparing the output mesh to a ground truth scan from our dataset. We select a sample of 10,000 points from each mesh, sampling uniformly over the surface area. We compare the nearest neighbor (NN) between each point in the two sampled point clouds, and evaluate two metrics: the NN distance, and the NN surface normal angular error.

We compare our method against a meshed COLMAP reconstruction, which is built from COLMAP's dense point cloud via Poisson surface reconstruction \cite{kazhdan2013screened}.

\section{Results}

\newcommand{\normImg}[1]{\includegraphics[width=0.16666\columnwidth]{#1}}
\newcommand{\normImgTrim}[1]{\includegraphics[trim={6.5cm, 8cm, 5.5cm, 8cm},clip, width=0.1666\columnwidth]{#1}}
\begin{figure}
    \centering
    \newcommand{\normRow}[1]{\normImg{#1/rgb.jpeg} & \normImg{#1/gt.png} & \normImg{#1/colmap.png} & \normImg{#1/iccv.png} & \normImg{#1/final_b5.png} & \normImg{#1/final_b5_alpha.png}}
    
    \newcommand{\normRowTrim}[1]{\normImgTrim{#1/rgb.jpeg} & \normImgTrim{#1/gt.png} & \normImgTrim{#1/colmap.png} & \normImgTrim{#1/iccv.png} & \normImgTrim{#1/final_b5.png}
    & \normImgTrim{#1/final_b5_alpha.png}}

    \begingroup
    \begin{tiny}
    \setlength{\tabcolsep}{0pt} 
    \renewcommand{\arraystretch}{0} 
    \begin{tabular}{cccccc}
             \normRow{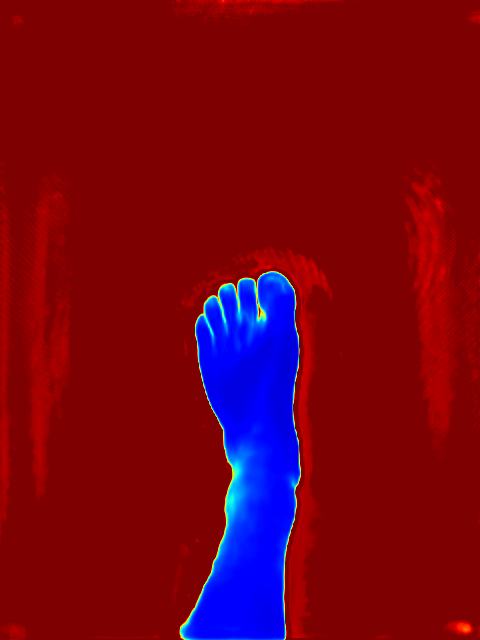} \\
             \normRow{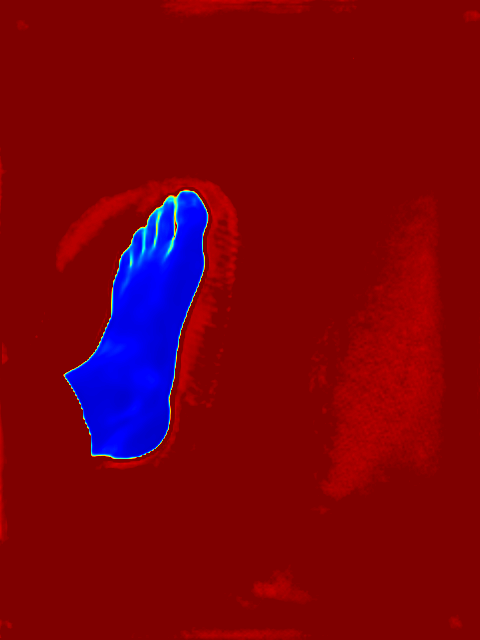} \\ 
             \normRow{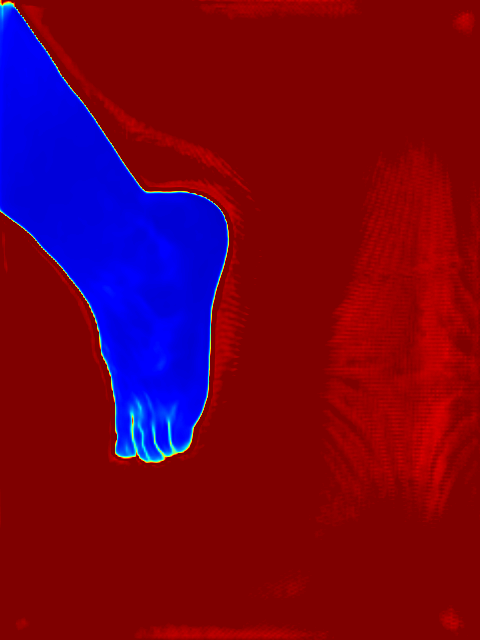} \\
             \normRow{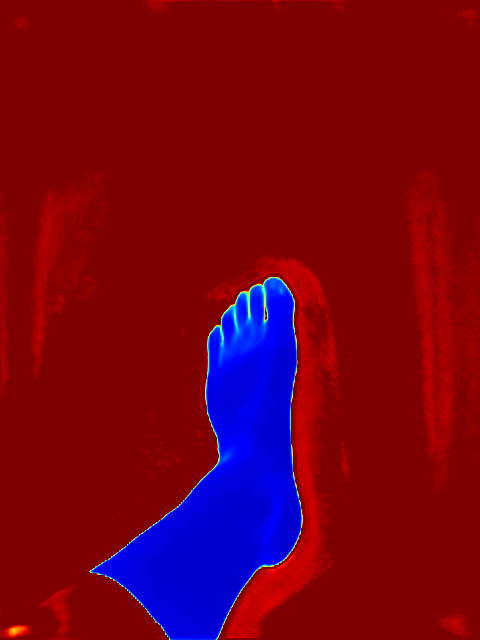}\\[3pt]
             \normRowTrim{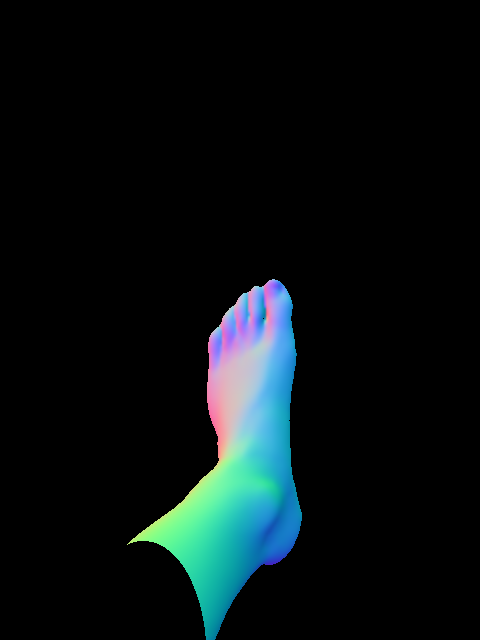}\\[3pt]

             RGB & GT & COLMAP \cite{MVS_2016_COLMAP} & Bae \etal \cite{SNfromRGB_2021_EESNU} & Ours & Ours - uncertainty \\[2pt]

            & & & & &  \begin{tabular}{lr}
            \multicolumn{2}{c}{\reflectbox{\includegraphics[width=0.1666\columnwidth]{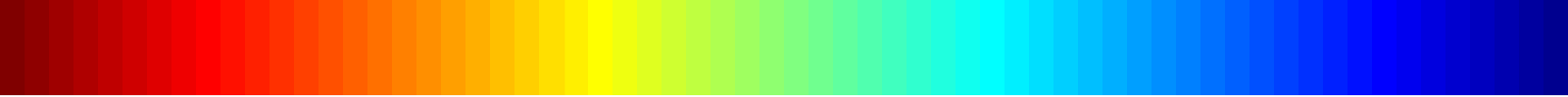}}} \\[2pt]
            $0^{\circ}$ & $60^{\circ}$
            \end{tabular}
    
    \end{tabular}
    \end{tiny}
    \endgroup
    \vspace{1mm}
    \caption{Qualitative comparisons of per-pixel normals, masking each prediction according to the GT. Further examples of in-the-wild surface normal predictions included in the supplementary.}
    \label{fig:normal_qual}
\end{figure}

\begin{table}[]
    \centering
    
    \begingroup
    \begin{footnotesize}
    \setlength{\tabcolsep}{4pt} 
    
\begin{standalone}

\renewrobustcmd{\bfseries}{\fontseries{b}\selectfont}
\newrobustcmd{\B}{\bfseries}

\begingroup
\setlength{\tabcolsep}{3pt} 

\begin{tabular}{c|*3{c}|*3{c}}
\toprule
\multirow{2}{*}{Method} & 
\multicolumn{3}{c|}{Normal error ($^{\circ}$) $\downarrow$} &
\multicolumn{3}{c}{Normal accuracy ($\%$) $\uparrow$} \\
& 
Mean & Median & RMSE & 
$11.25^{\circ}$ & $22.5^{\circ}$ & $30^{\circ}$
\\
\midrule
COLMAP \cite{MVS_2016_COLMAP} &
28.24 & 17.22 & 40.88 & 33.00 & 60.46 & 70.33 \\
Bae et al. \cite{SNfromRGB_2021_EESNU} &
28.54 & 25.89 & 32.79 & 12.43 & 41.27 & 59.75 \\
\hline
Ours {\scriptsize (aug. from \cite{SNfromRGB_2021_EESNU})} &
25.50 & 19.66 & 32.26 & 27.02 & 55.66 & 67.61 \\
Ours &
\B 11.30 & \B 9.24 & \B 14.43 & \B 62.06 & \B 91.73 & \B 96.21 \\
\bottomrule
\end{tabular}

\endgroup

\end{standalone}

    \end{footnotesize}
    \endgroup
    \vspace{1mm}
    \caption{Evaluation metrics of normal predictions, comparing against our baseline GT scans.}
    \label{tab:results_normals}
\end{table}

\paragraph{Normal prediction.} Table \ref{tab:results_normals} compares the accuracy of our surface normal predictor to the baseline methods. We outperform both methods across all metrics. When our predictor is trained with data augmentation from \cite{SNfromRGB_2021_EESNU}, the performance degrades substantially, suggesting that the proposed aggressive augmentation helps reduce the synthetic-to-real domain gap. Qualitative comparison in Figure \ref{fig:normal_qual} shows that our method captures the complex geometry around the toes.

\vspace{-3pt}

\begin{table}[]
    \centering
    
    \begingroup
    \begin{footnotesize}
    \setlength{\tabcolsep}{4pt} 
    
\begin{standalone}

\renewrobustcmd{\bfseries}{\fontseries{b}\selectfont}
\newrobustcmd{\B}{\bfseries}

\begin{tabular}{c}
\begin{tabular}{c|*3{c}|*3{c}}
    \toprule
     Method & \multicolumn{3}{c|}{NN chamfer error (mm) $\downarrow$} & \multicolumn{3}{c}{NN normal error ($^{\circ}$) $\downarrow$} \\
     & Mean & Median & RMSE & Mean & Median & RMSE \\
     \midrule
     Ours & 3.2 & 2.7 & \B 3.8     & \B 12.9 & \B 9.9 & \B 17.4 \\
     COLMAP & \B 2.7 & \B 1.8 & 3.9   & 20.3 & 13.7 & 27.7 \\ \bottomrule
\end{tabular}

\end{tabular}

\end{standalone}

    \end{footnotesize}
    \endgroup
    \vspace{1mm}
    \caption{3D results on fitting to all images available in each scan.}
    \label{tab:results_3d}
\end{table}

\begin{figure}
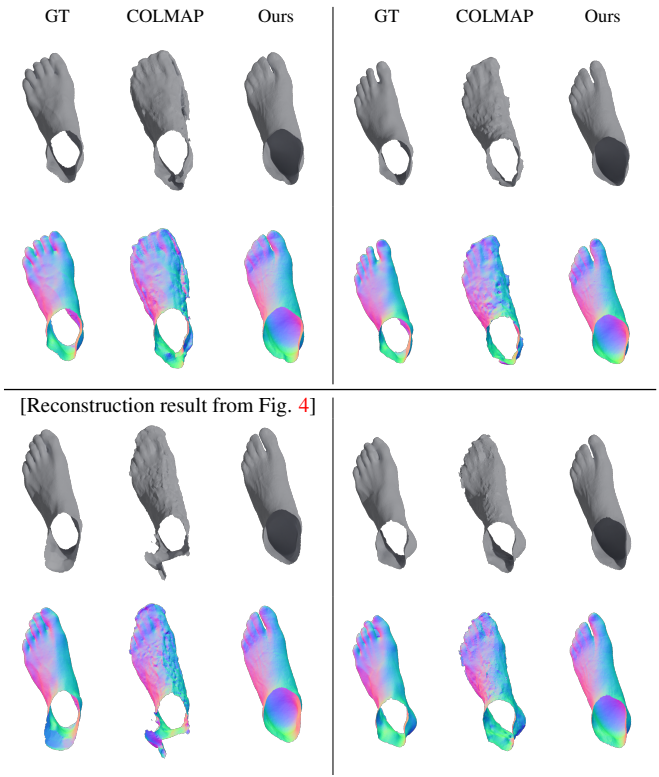

   \newcommand{\RImg}[1]{\includegraphics[width=0.12\columnwidth]{images/fits_3d/#1}}
   \newcommand{\RRow}[2]{\RImg{#1_GT/#2.png} & \RImg{#1_COLMAP/#2.png} & \RImg{#1_ours/#2.png}}
   \centering
  \begin{scriptsize}
   \begin{tabular}{ccc|ccc}
    GT & COLMAP & Ours & GT & COLMAP & Ours \\
      \RRow{0035}{rgb} & \RRow{0037}{rgb} \\
      \RRow{0035}{normals} & \RRow{0037}{normals}\\\midrule

        \multicolumn{3}{c|}{\footnotesize [Reconstruction result from Fig. \ref{fig:vis_reconstr}]} & \\
        \RRow{0038}{rgb} & \RRow{0039}{rgb} \\
      \RRow{0038}{normals} & \RRow{0039}{normals}
   \end{tabular}
   \end{scriptsize}

   \caption{Qualitative results on four reconstructions from all available images (approximately 30 per scan). We show geometry and surface normal renders of the ground truth 3D mesh, the COLMAP reconstruction, and the results of our optimization method. Further examples included in the supplementary.}
   \label{fig:reconstr}
\end{figure}

\paragraph{3D fitting results.} Table \ref{tab:results_3d} shows the quantitative performance of our method compared to COLMAP for the many view case. Qualitative examples of the reconstruction quality can be seen in Figure \ref{fig:reconstr}. The results suggest that COLMAP can capture some geometry more precisely (seen by the lower median chamfer error), at the cost of a noisy reconstruction that has some larger errors, and poor surface normal consistency. Conversely, our method obtains a smoother reconstruction with fewer outliers, and surface normals that much more closely resemble the ground truth.

\begin{figure}
    \centering
    \includegraphics[width=\columnwidth]{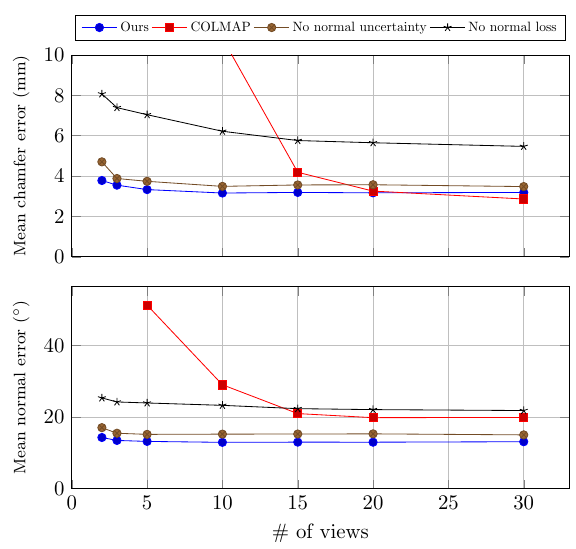}
    \caption{We show how the reconstruction quality of our method varies with number of input views, comparing with COLMAP, and with our method without normals and normal uncertainties. We show that COLMAP reconstruction fails for fewer than 15 views, whereas our method maintains accuracy for 3-5 views.}
    \label{fig:num_views}
\end{figure}

\paragraph{Number of views.} We show in Figure \ref{fig:num_views} our method's ability to perform accurate reconstructions with few available views. COLMAP achieves similar reconstruction accuracy for 30+ views, but fails almost completely for fewer than 15 views. Our method, on the other hand, is able to retain virtually all accuracy for as few as 3 views, and even handles the 2 view case with reasonable accuracy (better than COLMAP with 15).  Furthermore, the graph shows that the use of our normal loss both substantially improves the reconstruction quality, and significantly improves our method's ability to handle few views.

\begin{table}
   \centering
    \begingroup
    \begin{footnotesize}
    \setlength{\tabcolsep}{4pt} 
    
\begin{standalone}

\renewrobustcmd{\bfseries}{\fontseries{b}\selectfont}
\newrobustcmd{\B}{\bfseries}
\centering
\begin{tabular}{c}
\begin{tabular}{c|cc|cc}
      \toprule
       & \multicolumn{2}{c|}{\tiny {NN Chamfer error (mm) $\downarrow$}} & \multicolumn{2}{c}{\tiny NN Normal error ($^{\circ}$) $\downarrow$}\\
       & 3 view & 20 view & 3 view & 20 view \\
       \midrule
Ours & \B 3.41 &  \B 3.18 &  \B 13.58 & \B 13.12 \\
w/o keypoint uncertainty &  3.48 &  3.30 &  13.70 &  13.27 \\
w/o normal uncertainty &  3.86 &  3.57 &  15.57 &  15.37 \\
w/o normal loss &  6.39 &  5.64 &  24.23 &  22.13 \\

      \bottomrule
\end{tabular}

\end{tabular}

\end{standalone}

    \end{footnotesize}
    \endgroup
    \vspace{0.5mm}
   \caption{Ablation study on our 3D optimization method.}
   \label{table:ablation}

\end{table}

\vspace{-5pt}
\paragraph{Ablation study. } We show results of our ablation of our optimization process in Table \ref{table:ablation}. The data shows that, for both the few view and many view cases, our normal loss is crucial for accurate surface reconstruction. Furthermore, using normal uncertainty, and to a lesser extent keypoint uncertainty, further improve the quality of reconstruction.

\section{Conclusion}

\vspace{-3pt}

In this paper, we introduced our synthetic dataset \textit{SynFoot}, that effectively captures variation in foot images despite only containing 8 real world leg scans. We have developed a method for learning to predict normals with uncertainty purely from this synthetic data, using aggressive data augmentation to bridge the synthetic-to-real domain gap and obtain accurate surface normal predictions on real images. Experimental results show that this method significantly outperforms off-the-shelf normal predictors, and MVS reconstruction.  We have also shown a method for reconstructing a foot model from a small number of views that utilises the predicted normals with uncertainties, that outperforms typical MVS pipelines while requiring an order of magnitude fewer input images.

\section{Future work}

\begin{figure}
    \centering

    \begingroup

    \setlength{\tabcolsep}{2pt} 
    \newcolumntype{C}[1]{>{\centering\let\newline\\\arraybackslash\hspace{0pt}}m{#1}}

    \def\imH{0.455}

\settoheight{\imageheight}{%
  \includegraphics[width=\imH \columnwidth,keepaspectratio]{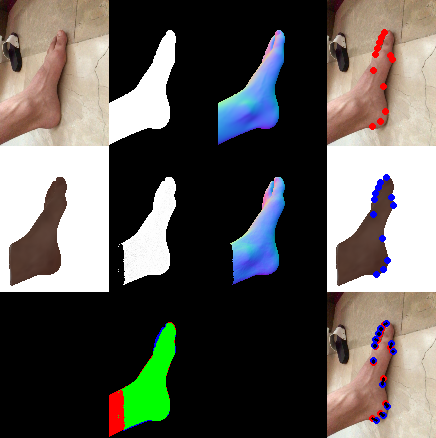}%
}

    \newcommand{\recheaders}{\footnotesize \begin{tabular}{*4{C{0.1\columnwidth}}}
    (a) & (b) & (c) & (d)
    \end{tabular} }

    \newcommand{\recrows}{\tiny \begin{tabular}{c}
    (i) \\[0.25\imageheight] (ii) \\[0.25\imageheight] (iii) 
    \end{tabular} }

    \begin{tabular}{p{0.25cm}cc}
    & \recheaders & \recheaders \\
    \recrows & \raisebox{-.5\height}{\includegraphics[width=\imH \columnwidth]{images/reconstr_vis/arkit_test_1/views/stage_01/view_00.png}} &
    \raisebox{-.5\height}{\includegraphics[width=\imH \columnwidth]{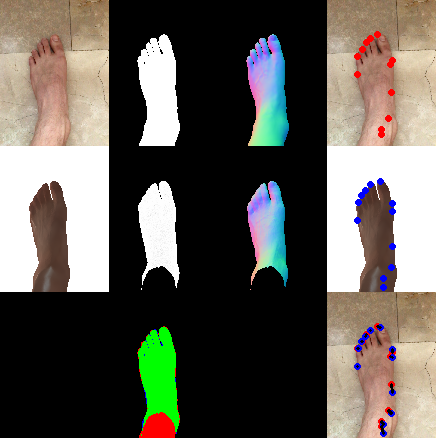}}
        
    \end{tabular}
    \endgroup
    \vspace{0.5mm}
    \caption{We show qualitative reconstruction on cameras obtained via AR tracking. We show here two out of the three used views, visualizing (a) RGB, (b) silhouette, (c) normals, and (d) keypoints, for (i) input image, (ii) fitted FIND model, and (iii) the error.}
    \label{fig:ar_reconstr}

\end{figure}

\label{sec:future_work}

\paragraph{Generalization. } While we only show results on the constrained case of a foot placed on the floor, nothing inherent in our method restricts us to this use case. As future work, we would like to extend this to posed feet in arbitrary positions, and to other body parts, such as hands.

\vspace{-10pt}

\paragraph{Cameras from AR tracking. } Our evaluation dataset shows results on cameras obtained via SfM, reliant on many views with significant overlap for accuracy. However, our method is also possible using camera parameters obtained via AR camera tracking. Although we do not have GT meshes, we show qualitatively on a private dataset with AR calibrated cameras that our optimization is  able to fit to real images successfully. Figure \ref{fig:ar_reconstr} shows one example of this, and further examples are included in the supplementary.

\section{Acknowledgements}
The authors acknowledge the collaboration and financial support of Trya Srl.


{\small
\bibliographystyle{ieee_fullname}
\bibliography{bib}
}

\pagebreak

\begin{standalone}
  \onecolumn
  \setcounter{section}{0}


\begin{center}
    \Large \textbf{\underline{Supplementary Material}}
\end{center}

\begin{figure}[!h]
    \centering
    \includegraphics[width=0.5\columnwidth]{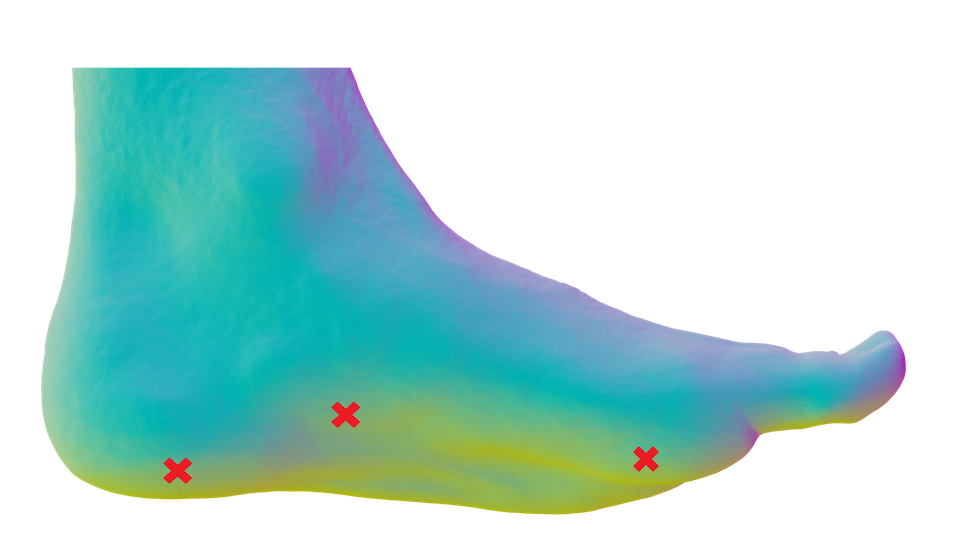}
    \caption{We show the `arch' keypoints labelled on an example foot, showing the surface normals.}
    \label{fig:kp_defs}
\end{figure}

\section{Keypoint definitions}

We define 12 keypoints on the foot:
\begin{itemize}
    \item \textbf{Toes} (5) - We label the most extremal point on each of the 5 toes
    \item \textbf{Width} (2) - We label the `inner extrema' and `outer extrema', the widest points on the front of the foot
    \item \textbf{Heel} (2) - We label two keypoints on the heel - one corresponding to the extremal (furthest `backwards') point at the bottom of the foot, and one at the point where this contacts the floor
    \item \textbf{Arch} (3) - We label three keypoints to define the arch of the foot - we achieve this by viewing the surface normals of the foot, and identifying the arch by the transition of colour from yellow (pointing downwards) to blue (towards the right of the foot) when viewed side on. As can be seen in Figure \ref{fig:kp_defs}, we label the three points defining the ends and highest point of the arch.
\end{itemize}

\section{Further examples}

\paragraph{Synthetic dataset.} In Figure \ref{fig:more_synth}, we show additional samples of our synthetic dataset.

\paragraph{In-the-wild performance.} In Figures \ref{fig:more_kp} and \ref{fig:more_normal}, we show further qualitative predictions on in-the-wild images, of our keypoint and normal predictors.

\paragraph{Reconstruction.} In Figure \ref{fig:more_qual3d}, we show further examples of our 3D reconstructions.

\paragraph{Comparison to Commercial. } We show in Figure \ref{fig:vs_sf2} a visual comparison of the quality of our reconstruction compared to a typical PCA commercial implementation, noting improved reconstruction quality around the toes.

\begin{figure}
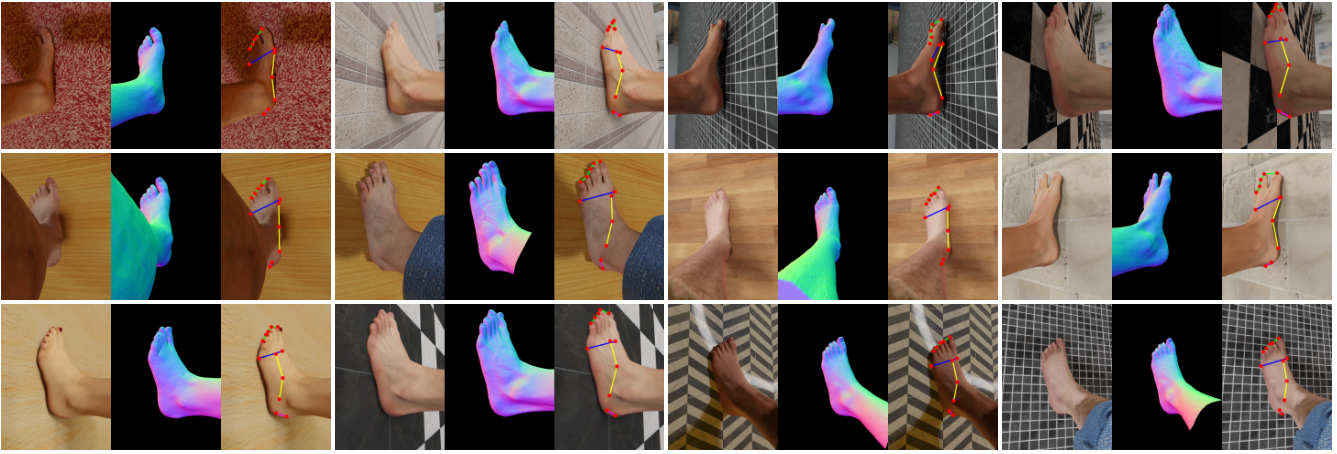

    \centering
    \newcommand\synth[1]{\includegraphics[width=0.25\columnwidth]{images/synth/000#1.png}}

    \begingroup
    \setlength{\tabcolsep}{1pt}
    \renewcommand\arraystretch{0.5}
    \begin{tabular}{cccc}
    \xintForfour #1#2#3#4 in {(003,009,015,017),
    (070,072,079,081),(135,139,168,169)}\do
    {%
        \synth{#1} & \synth{#2} & \synth{#3} & \synth{#4}\\
    }%
    &     
    \end{tabular}
    \endgroup

    \caption{Further samples from our synthetic dataset. We show (a) RGB, (b) normals and (c) keypoints}
    \label{fig:more_synth}
\end{figure}

\begin{figure}
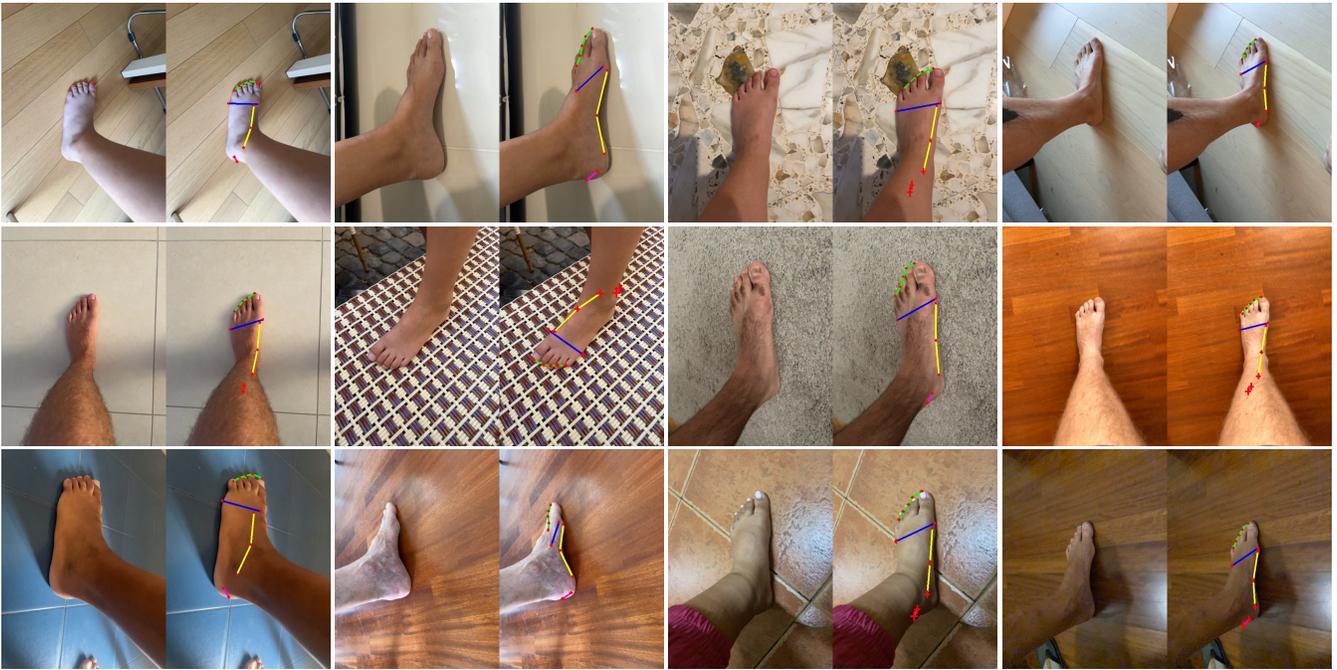

    \centering
    \newcommand\itwKP[1]{\includegraphics[width=0.25\columnwidth]{images/keypoint_itw/000#1.jpg}}
    \begingroup
    \setlength{\tabcolsep}{1pt}
    \renewcommand\arraystretch{0.5}
    \begin{tabular}{cccc}
    \xintForfour #1#2#3#4 in {(002,006,011,049),
    (081,113,165,245),(252,279,333,563)}\do
    {%
        \itwKP{#1} & \itwKP{#2} & \itwKP{#3} & \itwKP{#4}\\
    }%
    &
     
    \end{tabular}
    \endgroup
    \caption{Examples of our keypoint predictor on real, in-the-wild images.}
    \label{fig:more_kp}
\end{figure}

\begin{figure}
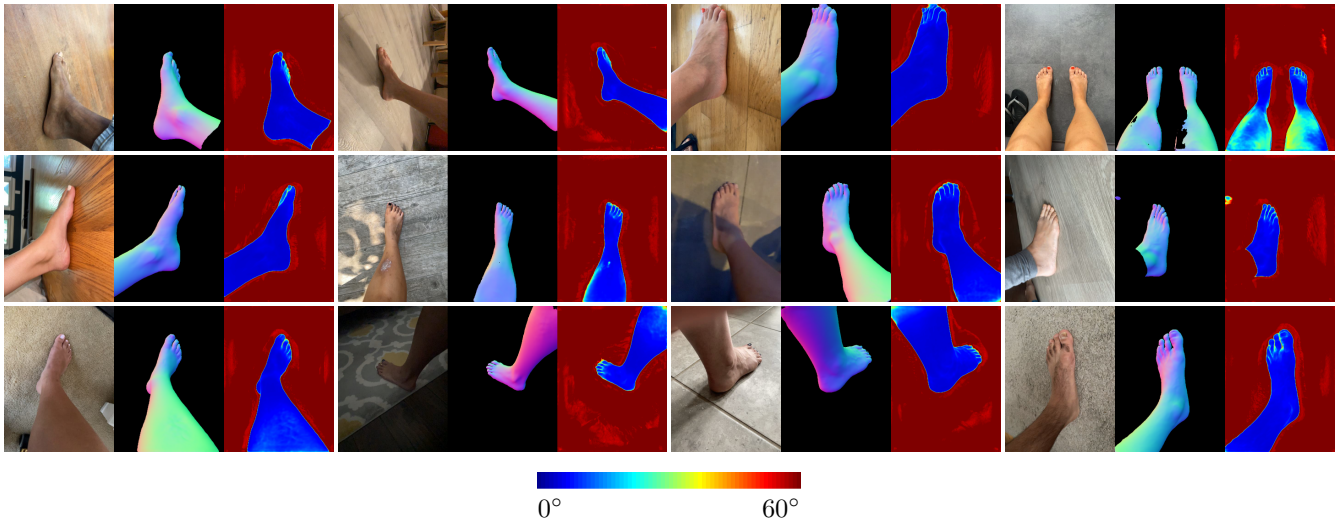

    \centering
    \newcommand\itwNorm[1]{\includegraphics[width=0.25\columnwidth]{images/normal_itw/foot_000#1_img.png}}

    \begingroup
    \setlength{\tabcolsep}{1pt}
    \renewcommand\arraystretch{0.5}
    \begin{tabular}{cccc}
    \xintForfour #1#2#3#4 in {(010,013,062,078),
    (091,095,098,097),(120,130,161,165)}\do
    {%
        \itwNorm{#1} & \itwNorm{#2} & \itwNorm{#3} & \itwNorm{#4}\\
    }%
    &\\[0mm]
    \multicolumn{4}{c}{\cbar{0.2\columnwidth}}
     
    \end{tabular}
    \endgroup

    \caption{Examples of our normal predictor on real, in-the-wild images in a variety of lighting conditions and viewpoints. For each, we show (a) RGB, (b) normals, (c) uncertainty. Note that our network is only trained on single foot images, but can still handle the multi-foot case.}
    \label{fig:more_normal}
\end{figure}

\begin{figure}
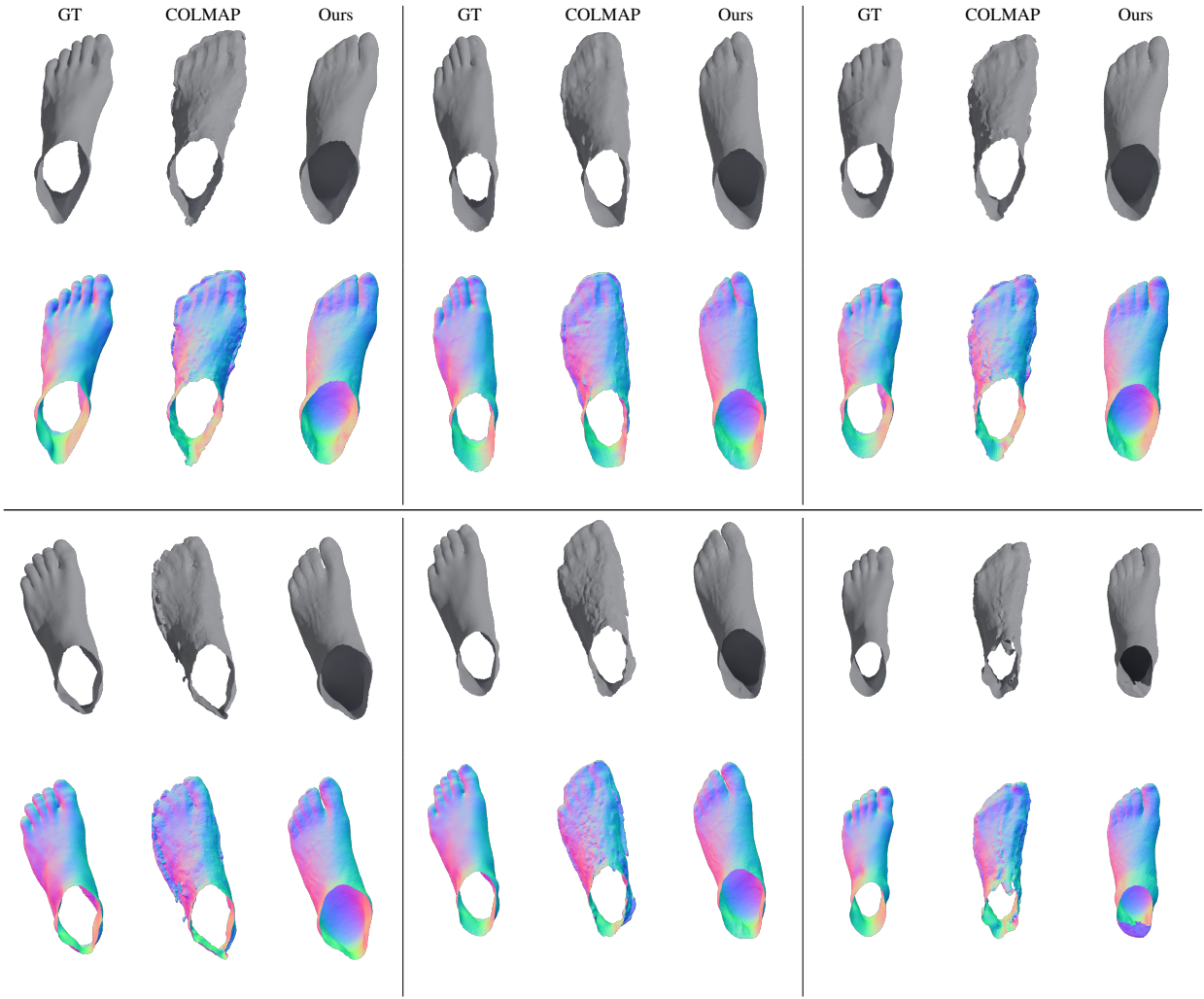

   \newcommand{\RImg}[1]{\includegraphics[width=0.08\columnwidth]{images/fits_3d/#1}}
   \newcommand{\RRow}[2]{\RImg{#1_GT/#2.png} & \RImg{#1_COLMAP/#2.png} & \RImg{#1_ours/#2.png}}
   \centering
  \begin{scriptsize}
   \begin{tabular}{ccc|ccc|ccc}
    GT & COLMAP & Ours & GT & COLMAP & Ours & GT & COLMAP & Ours \\
      \RRow{0044}{rgb} & \RRow{0041}{rgb} & \RRow{0048}{rgb} \\
      \RRow{0044}{normals} & \RRow{0041}{normals} & \RRow{0048}{normals}\\\midrule

        \RRow{0042}{rgb} & \RRow{0046}{rgb} & \RRow{0047}{rgb} \\
      \RRow{0042}{normals} & \RRow{0046}{normals} & \RRow{0047}{normals}
   \end{tabular}
   \end{scriptsize}

   \caption{Further qualitative results of reconstructions from our method, compared to COLMAP.}
   \label{fig:more_qual3d}
\end{figure}


\begin{figure}
    \centering
    \begingroup
    \setlength{\tabcolsep}{2pt} 
    
    \newcommand{\SFI}[1]{\includegraphics[width=0.15\columnwidth]{images/vs_sf2/#1.png}}
    \begin{tabular}{cc @{\hspace{0.1\columnwidth}} cc}
         \SFI{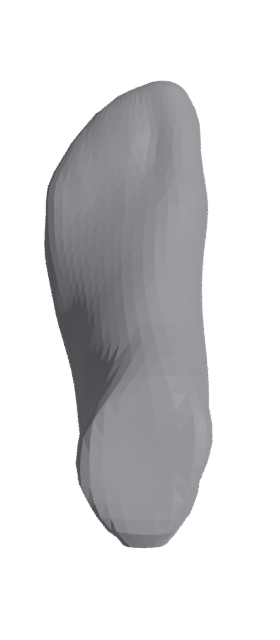} & \SFI{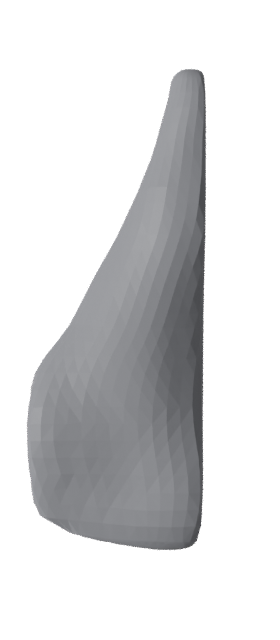} & \SFI{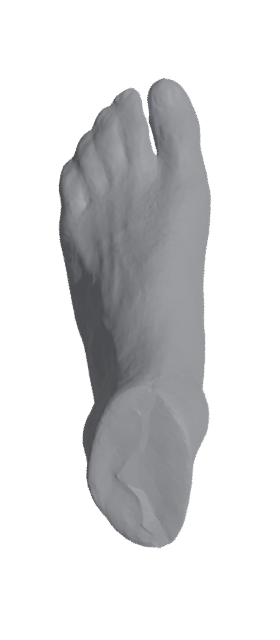} & \SFI{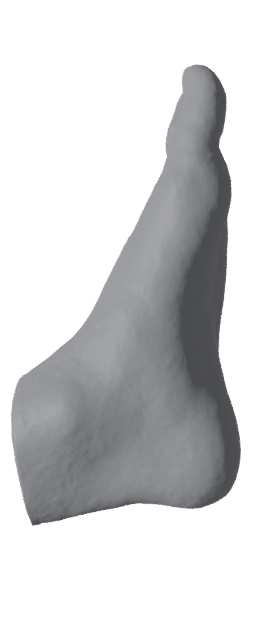}  \\
         \multicolumn{2}{c}{PCA commercial method} & \multicolumn{2}{c}{Ours}
    \end{tabular}
    \endgroup
    \vspace{0.5mm}    
    \caption{We compare a typical reconstruction of ours to that of a commercial, PCA based implementation.}
    \label{fig:vs_sf2}
\end{figure}

\section{Data augmentation details}

\noindent
For each augmentation, we use $p$ to denote the probability of applying it to a training image.

\paragraph{Downsample-and-upsample ($p=0.5$).} Bilinearly downsample the input image by ratio $r \in \mathcal{U}(0.2, 1.0)$ and upsample it back to its original resolution.

\paragraph{Horizontal flipping ($p=0.5$).} Horizontally flip the input image.

\paragraph{JPEG compression ($p=0.5$).} Compress the input \texttt{png} image into \texttt{jpeg}, with quality $q \in \mathcal{U}(10, 90)$.

\paragraph{Gaussian blur ($p=0.5$).} Apply Gaussian blur of kernel size $(7,7)$ and $\sigma \in \mathcal{U}(0.1, 10.0)$.

\paragraph{Gaussian noise ($p=0.5$).} Add Gaussian noise of $\sigma=0.01$ (the image is pre-normalized to $[0.0, 1.0]$).

\paragraph{Color jitter ($p=1.0$).} Apply \texttt{ColorJitter} augmentation in PyTorch with \texttt{brightness=0.5}, \texttt{contrast=0.5}, \texttt{saturation=0.5} and \texttt{hue=0.1}.

\paragraph{Grayscale ($p=0.02$).} Change the image into grayscale.

\paragraph{Perspective ($p=1.0$).} Rotate the camera around the yaw, pitch and roll axes, with angles $\theta_\text{yaw} \in \mathcal{U}(-20^\circ, +20^\circ)$, $\theta_\text{pitch} \in \mathcal{U}(-20^\circ, +20^\circ)$, and $\theta_\text{roll} \in \mathcal{U}(-180^\circ, +180^\circ)$.


\end{standalone}

\end{document}